%% file: emnlp-ijcnlp-2019.tex
\documentclass[11pt,a4paper]{article}
\usepackage[hyperref]{emnlp-ijcnlp-2019}
\usepackage{times}
\usepackage{latexsym}

\input{commands}

\aclfinalcopy 

\title{Knowledge Enhanced Attention \\for Robust Natural Language Inference}

\author{Alexander Hanbo Li \\
  Alexa AI, Amazon \\
  {\tt alexanderhanboli@gmail.com} \\\And
  Abhinav Sethy \\
  Alexa AI, Amazon \\
  {\tt sethya@amazon.com} \\}

\begin{document}
\maketitle
\begin{abstract}
Neural network models have been very successful at achieving high accuracy on natural language inference (NLI) tasks. However, as demonstrated in recent literature, when tested on some simple adversarial examples, most of the models suffer a significant drop in performance. This raises the concern about the robustness of NLI models. In this paper, we propose to make NLI models robust by incorporating external knowledge to the attention mechanism using a simple transformation. We apply the new attention to two popular types of NLI models: one is Transformer encoder, and the other is a decomposable model, and show that our method can significantly improve their robustness. Moreover, when combined with BERT pretraining, our method achieves the human-level performance on the adversarial SNLI data set.
\end{abstract}

\input{./subfiles/introduction}
\input{./subfiles/related_work}

\input{./subfiles/method}
\input{./subfiles/model}
\input{./subfiles/experiments}
\input{./subfiles/conclusion}

\bibliography{emnlp-ijcnlp-2019}
\bibliographystyle{acl_natbib}

\clearpage
\appendix
\input{./subfiles/supplement}

\end{document}

%% file: commands.tex
\usepackage{amsmath,bm,bbm,mathtools}
\usepackage[autostyle]{csquotes}
\usepackage{subcaption,multirow}
\usepackage{amssymb,pifont}
\usepackage{hyperref}       
\usepackage{url}            
\usepackage[scr=boondox]{mathalfa}
\usepackage{booktabs}       
\usepackage{amsfonts}       
\usepackage{nicefrac}       
\usepackage{microtype}      
\usepackage[colorinlistoftodos]{todonotes}
\usepackage{soul}
\usepackage{xcolor,colortbl}
\usepackage{booktabs}

\definecolor{anti-flashwhite}{rgb}{0.95, 0.95, 0.96}
\definecolor{antiquewhite}{rgb}{0.98, 0.92, 0.84}

\newcolumntype{g}{>{\columncolor{gray}}c}
\newcolumntype{w}{>{\columncolor{white}}c}
\newcolumntype{d}{>{\columncolor{anti-flashwhite}}c}
\newcolumntype{a}{>{\columncolor{antiquewhite}}c}

\newcommand{\cmark}{\ding{51}}%
\newcommand{\xmark}{\ding{55}}%
\newcommand{\diamondmark}{\ding{169}}%

\def\eqref#1{equation~\ref{#1}}

\def\1{\bm{1}}

\def\va{{\bm{a}}}
\def\vb{{\bm{b}}}

\def\vh{{\bm{h}}}

\def\vl{{\bm{l}}}

\def\vp{{\bm{p}}}

\def\vr{{\bm{r}}}

\def\vw{{\bm{w}}}

\DeclareMathAlphabet{\mathsfit}{\encodingdefault}{\sfdefault}{m}{sl}
\SetMathAlphabet{\mathsfit}{bold}{\encodingdefault}{\sfdefault}{bx}{n}



%% file: subfiles/introduction.tex
\section{Introduction}
\label{sec:introduction}
Natural language inference (NLI) \cite{bowman2015large} or recognizing textual entailment (RTE) \cite{dagan2013recognizing} is a task to predict whether a hypothesis sentence entails, contradicts or is neutral with a given premise sentence. Many public datasets have been constructed to help evaluate model performance on the NLI tasks, for example, the Stanford NLI (SNLI) \cite{bowman2015large}, the Multi-Genre NLI (MultiNLI) \cite{MultiNLI}, the SciTail dataset \cite{scitail} and etc, and some more recent and more difficult tasks like Swag \cite{zellers2018swag} and HellaSwag \cite{zellers2019hellaswag} for commonsense inference.

One line of work for NLI is to get sentence embeddings for both the premise and hypothesis \cite{bowman2015large,bowman2016fast,nie2017shortcut,shen2018reinforced}. The performance of these models rely on the information retained in the sentence embeddings. Another type of models make use of token level interaction. These methods align each token in the premise with similar tokens in the hypothesis, and vice versa. The major differences within this type of models are how the tokens are aligned and how many times the alignments are calculated. In \citet{rocktaschel2015reasoning,chen2016enhanced,chen2017external}, there are two separate networks, one for premise and the other for hypothesis. For each token in the premise(hypothesis), its alignment in the hypothesis(premise) is only calculated for once at a ``cross attention" stage. Before and after this stage, a word can only attend to the other words in the same sentence. Transformer models like BERT \cite{bert} and OpenAI Transformer \cite{radford2018improving} instead read the concatenation of the premise and hypothesis as one sentence (with special classification [CLS] and separation tokens [SEP]). Therefore, the attentions between words in premise and hypothesis are calculated for multiple times, depending on the number of layers.

However, \citet{glockner2018breaking} points out that the state-of-the-art NLI models are limited in their robustness and generalization ability. One reason is that large datasets like SNLI can be homogeneous, and the sentences may also have annotation artifacts \cite{gururangan2018annotation}. The authors hence create an adversarial NLI test set with examples that capture various kinds of lexical knowledge. The premises are taken from SNLI training set, and hypotheses are then constructed by replacing a single word or phrase by its lexical related word (phrase). Entailment examples are generated by replacing a word with its synonym or hypernym, contradiction examples are by replacing a word with its mutually exclusive co-hyponyms and antonyms, and neutral examples are by replacing a word with its hyponyms.



It turns out that many aforementioned models that achieve reasonable results on clean test set perform much worse on the adversarial data, with more than $20\%$ accuracy drop. For example, ESIM \cite{chen2016enhanced} achieves 87.9\% accuracy on the clean SNLI test set, but only gets 65.6\% on the adversarial SNLI data. Despite the new task being considerably simpler, the drop in performance is substantial. 

In this paper, we investigate the attention mechanism that is widely used across many models. We link the multi-head attention to structured embeddings, and show that one can directly add diverse external knowledge to a multi-head attention. This modification is straightforward, and does not require extra parameters. Any model that has attention components can benefit from this method. To showcase it, we apply the method to two kinds of model structures: one is decomposable models like \citet{parikh2016decomposable,chen2016enhanced}, and the other is Transformer models \cite{vaswani2017attention,bert,radford2018improving}.

Since adding external knowledge to multi-heads essentially allows a model to explore beyond the data distribution of a specific task, the pretraining procedure should also be able to robustify a model. Therefore, we also use BERT \cite{bert} and OpenAI pretraining \cite{radford2018improving} and investigate how they alone or combined with our proposed method can further improve the model robustness.

%% file: subfiles/related_work.tex
\section{Related Work}
\label{sec:related}
\citet{chen2017external} explored adding external knowledge to make NLI inference more robust. The Wordnet knowledge is summarized as relational feature vectors, which are used in three components, including word alignment, inference composition, and calculation of word weights for the weighted pooling. As a result, the methods are bound to the KIM model. Some non-neural NLI models also incorporate external knowledge graphs by model-specific engineering \cite{raina2005robust,haghighi2005robust,silva2018recognizing}. But each method is still only designed for one model. In fact, incorporating external rules in a deep learning model generally requires substantial model architecture adaption. In contrast, our method can be treated as a component which can be added to any model that uses attention mechanism.

\citet{zhong2018improving} focused on science question domain, and proposed to use knowledge graph from external resources like ConceptNet and DBpedia to help NLI, by retrieving relevant information related to the premise and hypothesis. \citet{kang2018adventure} proposed a GAN-type model to generate adversarial examples guided by knowledge for robust training. \citet{hu2016harnessing} proposed to encapsulate logical rules into a neural network by forcing the network to emulate a rule-based teacher. Other than directly improving the model, \citet{faruqui2014retrofitting} and \citet{mrkvsic2016counter} retro-fit or counter-fit the word embeddings to linguistic resources. These embedddings are model-agnostic and can be used in many tasks. Our method instead does not require any extra parameters or extra training, but directly incorporate the knowledge into the attention function.

%% file: subfiles/method.tex
\section{Method}
\label{sec:method}
Multi-head attention plays an essential role in the Transformer based models. To decide the attention of one word on the others, one needs to define a proper similarity measure. In the Transformer, the similarity between two word embeddings $\vw_1$ and $\vw_2 \in \mathbb{R}^d$ are defined as the inner product
\begin{align}\label{eq:word_sim}
    (L\vw_1)^\top (R\vw_2),
\end{align}
where $L, R \in \mathbb{R}^{d_k \times d}$ are two transforming matrices, mapping vectors from the original embedding space to others. In fact, the transforming matrices $L$ and $R$ are strongly related to the structured embedding \cite{bordes2011learning} in graph models.

\subsection{Structured Embedding}
\label{ssec:se}
In \citet{bordes2011learning}, the knowledge bases are considered as graph models, in which each individual node stands for an element of the database, and each edge defines a relation between entities. For example, in Wordnet, entities are words and edges are lexical relations like hypernymy, synonymy, antonymy and etc. Each entity is represented by a $d$-dimensional dense vector (word embedding) $\vw \in \mathbb{R}^k$. And within the embedding space, there is a specific similarity measure that captures the relation between entities. This potentially asymmetric relation (denoted by $\vr$) can be modeled by two transformations $f_L$ and $f_R$, which together define a similarity measure
\begin{align}\label{eq:se_sim}
    S(\vw_i, \vw_j) = d(f_L (\vw_1), f_R (\vw_2)).
\end{align}
For example, $d(\va, \vb) = \|\va - \vb\|^{-1}_p$, $d(\va, \vb) = \cos(\va, \vb)$, or $d(\va, \vb) = \va^\top \vb$. The goal is to find a similarity function such that $\vw_i$ and $\vw_j$ are more likely to have the relation $\vr$ if they are closer in terms of the similarity measure in the transformed space.

For example, in Wordnet, the relation $\vr$ can be \textit{\_is\_antonym\_of\_}. Two words that are antonyms may not be close in the original embedding space, but will be close in some antonym embedding space, defined by the transformations $f_L$ and $f_R$. In another word, if using (\ref{eq:word_sim}) to be the similarity measure, then two antonyms $\vw_i$ and $\vw_j$ can have small similarity $\vw_i^\top \vw_j$, but large similarity $(f_L\vw_i)^\top (f_R\vw_j)$.

Therefore, if we can learn a pair of transformations $(f_L, f_R)$ for each lexical relation, then in each transformed space, two words with the corresponding relation are more likely to be aligned together.

\subsection{Multi-head Attention}
Multi-head attention \cite{vaswani2017attention} is closely related to structured embedding. Each head linearly projects the query and key into a new embedding space, and compare them therein. In this case, $f_L = L$ and $f_R = R$ are both linear.

Using the same notation in \citet{vaswani2017attention}, the queries, keys and values are packed into rows of $Q$, $K$, $V$, and the attention function is defined as
\begin{align}\label{eq:attention}
    \text{Attention}(Q, K, V) = \text{softmax}\left( \frac{Q K^\top}{\sqrt{d_k}} \right) V.
\end{align}
Then the multi-head attention is
\begin{align}\label{eq:multi-head}
    \text{MH}(Q, K, V) = \text{Concat}(\text{head}_1, \ldots, \text{head}_h) W^o
\end{align}
where $\text{head}_i = \text{Attention}(Q L_i, K R_i, V W_i)$. If $(L_i, R_i)$ successfully learns a lexical information, say hyponym, then for each query word, the $i$-th $\text{head}_i$ will carry more information of its hyponyms in the keys and values. 

The softmax part of (\ref{eq:attention}) determines which rows of $V$ that each query should attend to. If the matrices $L_i$ and $R_i$ are only learned through a specific task, then they might learn some unique pattern that only occurs in this task. Pretraining is an effective method to solve this problem and to learn more robust structured embeddings, but they require extra learning of large-scale data sets.

In the case that we know what kind of robustness we want (e.g. lexical) and prefer more interpretability of the model, we show that we could directly modify the behavior of the multi-head attention to incorporate the external knowledge. For example, if lexical robustness is of interest, we can add a different lexical relation to each head of (\ref{eq:multi-head}). Then the multi-heads altogether will carry all the lexical information, and hence be more robust to lexical noise.

\subsection{Direct Modification of the Multi-heads}
\label{ssec:direct}
For each head $i$, if $L_i$ and $R_i$ represent one lexical relation, then any two words $w_1$, $w_2$ having this relation will be closer in the transformed space,\footnote{From now on, we will use $w_i$ to represent a word token, and the bold letter $\vw_i$ to represent the corresponding word embedding.} meaning that $(L_i \vw_1)^\top (R_i \vw_2)$ will be larger. While such linear mappings $L_i$ and $R_i$ can be hard to learn for some concepts, we can always directly change the similarity they produce. Therefore, we propose to add an offset $b > 0$ to the similarity whenever two words have the relation of interest. More formally\footnote{$(w_1, w_2) \in \vr$ means $w_1$ has relation $\vr$ with $w_2$.}, for head $i$,
\begin{align*}
    &\text{Similarity}_i(\vw_1, \vw_2)  \\
    =&\begin{cases}
    (L_i\vw_1)^\top (R_i\vw_2) + b, \text{ if } (w_1, w_2) \in \vr_i, \\
    (L_i\vw_1)^\top (R_i\vw_2), \text{ otherwise.}
    \end{cases}
\end{align*}
Or in the matrix form,
\begin{align}\label{eq:attention_b}
    \text{head}_i = \text{softmax}\left( \frac{(QL_i) (KR_i)^\top}{\sqrt{d_k}} + b B_i \right) VW_i,
\end{align}
where $B_{i}[p,q] = 1 > 0$ if $(w_p, w_q) \in \vr_i$, and $B_{i}[p,q] = 0$ otherwise. The magnitude of hyperparameter $b$ controls the attention weights, and can be tuned on a validation set. If $b \to \infty$, then each word in the query will only attend to the words in the key that have relation $\vr_i$ with it.

Each matrix $B_i$ represents one lexical relation, which is then incorporated into head $i$ using (\ref{eq:attention_b}). With all the heads combined together, the multi-head attention (\ref{eq:multi-head}) then carries all the lexical information we incorporated, and hence makes the model more robust to lexical noise.

Note that in the current layer, even $L_i$ and $R_i$ may not be impacted by $B_i$ directly, the subsequent layers will learn to use the information $B_i$ carries, and hence change their behavior accordingly. We will further investigate it in the experiment section.

\subsection{Pretraining and Robustness}
In the previous section, we show how to add external knowledge into the attention function. The layers will learn to use $B_i$'s but not to predict them. Pretraining of a deep network, on the other hand, figures out some of the relations in an unsupervised way. The tasks can be predicting the next sentence, predicting masked words and etc \cite{bert,radford2018improving}. Pretraining \cite{bert,radford2018improving,peters2017semi,peters2018deep,dai2015semi,howard2018universal} could potentially learn rich knowledge including lexical ones, and has already been shown to improve the model performance on many NLP tasks \cite{wang2018glue}. On the NLI tasks, \citet{bert,radford2018improving} improve the accuracy absolutely by about 10\% on MultiNLI, and about 2\% on SNLI.

While in some cases, pretraining may only bring marginal improvement, we show later that they can have significant impact on the model robustness to adversarial data. The reason is that if the model is only trained for one specific task, it may catch some patterns that are only valid for this task, but pretraining forces the model to learn diverse structured embeddings (\ref{eq:word_sim}).

We also apply our method to BERT when we fine-tune it for the SNLI task, and show that even on this complicated model, our method can still further improve the robustness. On the adversarial SNLI data, BERT with our method achieves an accuracy of 94.1\%, which is equal to estimated human performance. More details are in the experiment section.

%% file: subfiles/model.tex
\section{Model Architecture}
\label{sec:model}
Multi-head attention can be applied to various kinds of models, and hence our proposed method in section \ref{ssec:direct} can also be widely used. In this section, we focus on two types of models that achieve the state-of-the-art results on NLI tasks. The first model follows the structure of the decomposable models \cite{parikh2016decomposable,chen2016enhanced,chen2017external,tay2017compare}, but uses multi-head attention. It consists of two neural networks - one for premise and the other for hypothesis. The second model is a Transformer encoder followed by a classification layer. The premise and hypothesis sentences are concatenated and fed to the encoder.

\subsection{Model I: Decomposable Model}
\label{ssec:decomp_model}
\begin{figure*}[tb]
\centering
\includegraphics[width=0.90\linewidth]{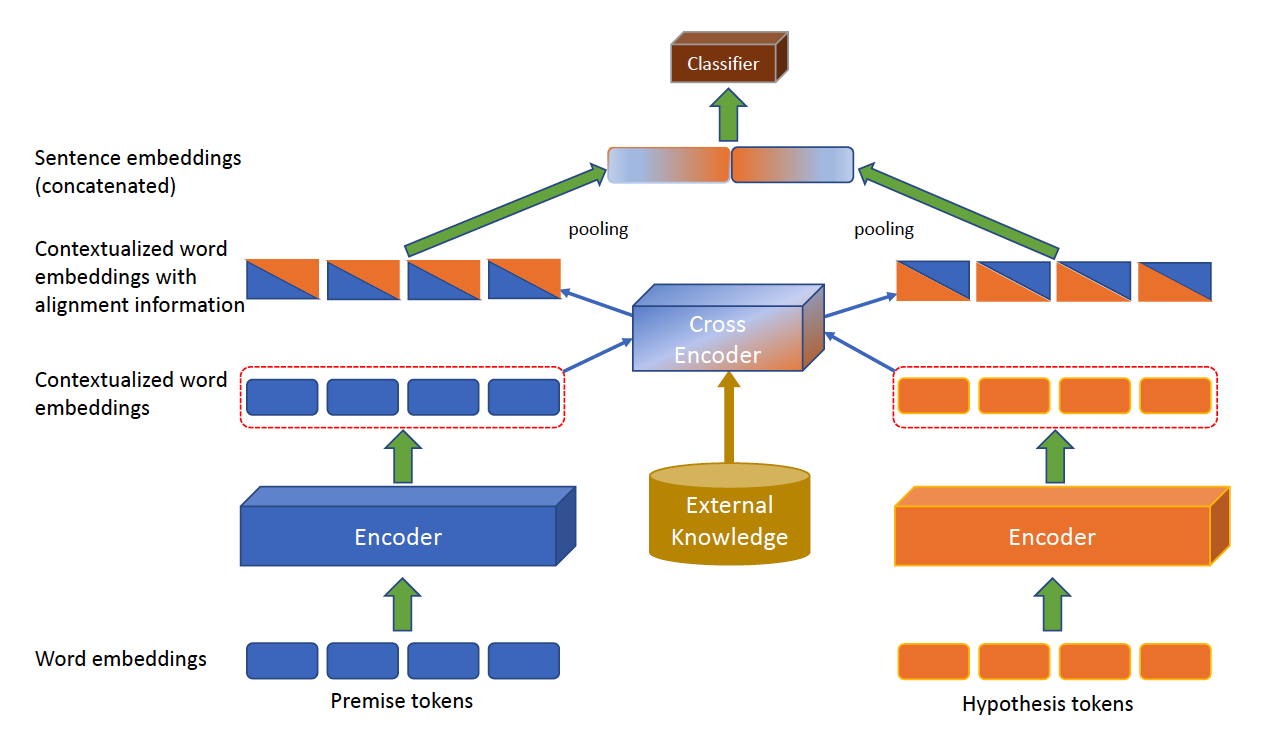}
\caption{The architecture of the decomposable model I.}
\label{fig:decomp_model}
\end{figure*}

The key components of Model I are Transformer encoder and cross-encoder. The visualization of model I is in Figure \ref{fig:decomp_model}.

\paragraph{Encoder:}
The first step is to get the contextualized word vectors using Transformer encoders with multi-head self-attention. We denote the encoders for premise and hypotheses as $E_p$ and $E_h$ respectively. Each encoder is a stack of $N$ encoder layers, and each layer has the same structure and residual connection as the encoder layer in \citet{vaswani2017attention} (Figure 1 therein).

We denote $\bm{P} = (\vp_1, \ldots, \vp_{l_1})$ and $\bm{H} = (\vh_1, \ldots, \vl_{l_2})$ to be the sequence of word embeddings in the premise and hypothesis respectively. Each word embedding also has its positional encoding added. The contexualized words are then
\begin{align*}
    &\bm{P}^{context} = E_p(\bm{P}) = (\vp^c_1, \ldots, \vp^c_{l_1}), \\
    &\bm{H}^{context} = E_h(\bm{H}) = (\vh^c_1, \ldots, \vh^c_{l_2}).
\end{align*}

\paragraph{Cross-encoder:}
The second step is to calculate the cross-attentions between the premise and hypotheses. The cross-encoder is a stack of $N_{cr}$ layers, and each layer follows the same structure of the decoder layer in the Transformer \cite{vaswani2017attention} (Figure 1 therein). Denote the cross-encoders as $E^{cr}_p$ and $E^{cr}_h$, we have new word vectors
\begin{align*}
    \bm{P}^{cross} &= E^{cr}_p(\bm{P}^{context}, \bm{H}^{context}, \bm{B}) \\
    &= (\vp^{cr}_1, \ldots, \vp^{cr}_{l_1}), \\
    \bm{H}^{cross} &= E^{cr}_h(\bm{H}^{context}, \bm{P}^{context}, \bm{B}) \\
    &= (\vh^{cr}_1, \ldots, \vh^{cr}_{l_2}).
\end{align*}
Here, $\bm{B}$ is the stack of matrices of external information, which we can choose to add to the multi-head attentions in the cross-encoder layers. The output word embeddings carry both contextualized and alignment information. 

\paragraph{Classifier}
We then get sentence embeddings by max and average pooling on the cross-encoder outputs, that is, $\vp^{sent} = [\max_i (\vp^{cr}_{i}); \text{mean}_i (\vp^{cr}_{i})]$, and $\vh^{sent} = [\max_i (\vh^{cr}_{i}); \text{mean}_i (\vh^{cr}_{i})]$\footnote{``;" means concatenation.}. Finally the concatenation of two vectors $[\vp^{sent}; \vh^{sent}]$ is fed into a classifier to predict the class label.

\subsection{Model II: Transformer Encoder}
The previous model consists of two different encoders for premises and hypothesis separately. In this section, we follow the model design of BERT which only uses the Transformer encoder. The input to model II is the concatenation of the premise and hypothesis, separated and appended by special tokens. For example, if the premise sentence is ``\textit{A cat is sleeping under the couch.}" and the hypotheses is ``\textit{There is an animal present.}", then the tokens of the concatenated sentence are: \textit{[CLS] A cat is sleeping under the couch . [SEP] There is an animal present . [SEP]}

Here, [CLS] stands for the task ``classification". Following BERT, besides adding the positional embeddings to all word vectors, we also add the segment embeddings representing sentence types (premise or hypotheses).

The workflow of Model II is as follows. Word embeddings are passed into a Transformer encoder with $N$ encoder layers. After the final encoder layer, we use the embedding of the [CLS] token as the pooling of all word vectors, and this embedding is then fed into a classifier to get the final prediction.

%% file: subfiles/experiments.tex
\section{Experiments}
\label{sec:experiments}
In order to showcase how the proposed method makes NLI models more robust, we train different models on two NLI datasets -- SNLI \cite{bowman2015large} and MultiNLI \cite{MultiNLI}. We consider two scenario: in the first one, we train models on SNLI and test them on both clean and adversarial SNLI data \cite{glockner2018breaking}; and in the second scenario, we train on the combination of SNLI and MNLI, and test on clean MNLI, clean SNLI, and adversarial SNLI data. In both cases, we report the accuracy of models. On the adversarial daat, we also report the absolute increase of accuracy by using the proposed method, and the precision and recall of each class. Finally, we also link the proposed ``quick fix" method to pretraining, and compare our results with BERT and OpenAI GPT. 

\subsection{Model Details}
\label{ssec:model_architecture}

\paragraph{Model I:}
For the decomposable Model I, we denote the number of encoder layers as $N$, the number of cross-encoder layers as $N_{cr}$, and the number of multi-heads as $A$. All sub-layers and the embedding layers have hidden size $d_{model}$. Throughtout the experiments, we fix $d_{model} = 300$ and $A =  5$ for model I, so each head has dimension $d_{model}/A = 60$. Both the encoder and cross-encoder contain a position-wise feed-forward network,
\begin{align*}
    \text{FFN}(x) = \text{gelu}(xW_1 + b_1)W_2 + b_2,
\end{align*}
where gelu is a Gaussian error linear unit proposed in \citet{hendrycks2016gaussian}. The hidden layer size of FFN is $d_{ff} = 512$. The classifier reads in the concatenation of $\vp^{sent}$ and $\vh^{sent}$, and consists of two layers: $\text{Classifier}(x) = \text{tanh}(xW_1 + b_1)W_2 + b_2$.
The input size and hidden layer size are all $4 d_{model} = 1200$, and the output size is 3. The word embeddings are initialized using Glove embeddings \cite{pennington2014glove} trained on common crawl (840B tokens, 2.2M vocab, cased).

\paragraph{Model II:}
Since the input is concatenated premise and hypotheses, Model II only contains an encoder and a classifier. We again denote the number of encoder layers as $N$. Following the same notations in model I, we set $d_{model}=300$ and $A = 5$. The feed-forward network has the same structure, but with $d_{ff} = 4 d_{model} = 1200$. The classifier is the same as in Model I. To further investigate the impact of our proposed method, we random initialize the word embeddings.

\paragraph{BERT and other parameters}
The positional and segment embeddings are learned jointly with the models. All the weights in linear layers and embeddings are initialized from Gaussian distribution $\mathcal{N}(0, 0.02^2)$, and the biases are initialized to be zero. We use the BERT version of Adam optimizer \cite{kingma2014adam} with warm-up percentage $10\%$ and learning rate $0.0001$, and all the other parameters set to be default. We use cross-entropy loss for the classification. For both model I and II, the number of epochs is 5. For fine-tuning BERT, the number of epochs is 3, and we use the PyTorch implementation of BERT (\url{https://github.com/huggingface/pytorch-pretrained-BERT}) with all the parameters set to default.

\subsection{Wordnet Knowledge}
WordNet \cite{miller1995wordnet} is a large lexical database, where English words are grouped into different sets, each expressing a distinct concept. We focus on five concepts: synonym, hypernym, hyponym, antonym and co-hyponym. In summarization, there are in total 753086 hypernymy, 753086 hyponymy, 3674700 co-hyponymy, 6617 antonymy and 237937 synonymy.


\paragraph{Wordnet baseline}
The adversarial SNLI dataset contains 7164 contradiction, 982 entailment, and 47 neutral instances. The Wordnet baseline on the data set is \textbf{85.8\%} accuracy \cite{glockner2018breaking}. Denote the original word as $w_p$ and the replaced word as $w_h$, the method is to predict entailment if $w_p$ is a hyponym of $w_h$ or if they are synonyms, neutral if $w_p$ is a hypernym of $w_h$, and contradiction if $w_p$ and $w_h$ are antonyms or co-hyponyms.

\paragraph{Adding Wordnet knowledge to multi-heads}
For model I and II, since we use 5 concepts and also 5 heads, each concept will be added to a different head. For fine-tuning BERT, the number of heads is 12, and we only add the concepts to the first 5 heads, and leave the rest unchanged. And for Model II and BERT, the Wordnet knowledge is only added to a pair of words when they belong to different segments (i.e. sentence types).

\subsection{Analysis of Layer Modification}
\paragraph{Model I:}
We test the effectiveness of method \ref{ssec:direct} by adding the external knowledge to different cross-encoder layers, and compare the accuracy on both clean and adversarial data. We fix $N=1$, and compare the settings of $N_{cr} = 1, 2$. The SNLI result is summarized in Table \ref{tab:model_I_direct}, and the MNLI result is in Appendix (Table \ref{tab:model_I_direct_mnli}).

\begin{table*}[htb]
\small
\centering
\caption{Accuracy of \textbf{Model I} and \textbf{Model II} trained on \textbf{SNLI} and tested on both clean and adversarial SNLI data. We set the parameter $b=10$. The numbers in the modified layers are the layer indices where we add external knowledge using method \ref{ssec:direct}. The absolute increase is the improvement of accuracy compared to the corresponding baseline model.}
\label{tab:model_I_direct}
\vspace{0.10in}
\begin{tabular}{lccccdad}
\toprule
\textbf{Model} & \multirow{2}{*}{$\textbf{N}$} & \multirow{2}{*}{$\textbf{N}_{cr}$} & \textbf{Modified} & \textbf{No. of} & \textbf{Accuracy} & \textbf{Accuracy} & \textbf{Absolute}\\
\textbf{type} & & & \textbf{layer(s)} & \textbf{parameters} & \textbf{(clean)} & \textbf{(adversarial)} & \textbf{increase} \\
\toprule
\multirow{6}{*}{Model I} & 1 & 1 & none & 6.8m & 86.0\%          & 50.3\%  & baseline \\
\cmidrule{2-8}
 & 1 & 1 & 1    & 6.8m & 86.3\%          & 71.1\%  & 20.8\% \\
\cmidrule{2-8}
 & 1 & 2 & none & 8.9m & 87.2\%          & 54.1\%  & baseline \\
\cmidrule{2-8}
 & 1 & 2 & 1    & 8.9m & \textbf{88.5\%} & \textbf{79.4\%} & 25.3\% \\
\cmidrule{2-8}
 & 1 & 2 & 2    & 8.9m & 87.9\%          & 77.3\%  & 23.2\% \\
\cmidrule{2-8}
 & 1 & 2 & 1,2  & 8.9m & 88.3\%          & 78.6\%  & 25.5\% \\

\toprule
\multirow{5}{*}{Model II} & 3 & - & none   & 11.0m & 81.5\% & 40.9\% & baseline \\
\cmidrule{2-8}
 & 3 & - & 1      & 11.0m & 82.5\% & 55.4\% & 14.5\% \\
\cmidrule{2-8}
 & 3 & - & 2      & 11.0m & 82.2\% & 54.8\% & 13.9\% \\
\cmidrule{2-8}
 & 3 & - & 3      & 11.0m & 81.7\% & 53.9\% & 13.0\% \\
\cmidrule{2-8}
 & 3 & - & 1,2,3  & 11.0m & 82.4\% & 57.6\% & 16.7\% \\
\bottomrule
\end{tabular}
\end{table*}

From the tables, we observe a significant boost of accuracy on adversarial data if we add the external knowledge to some of the cross-encoder layers. For example, on SNLI, the accuracy of vanilla model I ($N=1$, $N_{cr}=1$) drops from 86\% on clean data to 50.3\% on the adversarial data. By adding external knowledge to the cross-encoder, the accuracy on adversarial data quickly imporves to 71.1\%, with an absolute improvement of 20.8\%. On the clean data, the external knowledge also brings about 1\% improvement.

Another observation is that our proposed method \ref{ssec:direct} works the best when applied to the consecutive cross-encoder layers including the first one. We think one reason is that each word will carry some noisy lexical information from the other sentence after the first cross-encoder layer. So if the model only gets the lexical relation from the second layer, it will receive a weaker signal compared to getting the information at the very beginning.


\paragraph{Model II:}
For Model II, we compare three settings $L=1,2,3$, and add external knowledge to some of the layers. The SNLI result is in Table \ref{tab:model_I_direct}, and the MNLI result is in Appendix (Table \ref{tab:model_II_direct_mnli}). On the adversarial data, our proposed method still provides about 15\% improvement when trained on SNLI, and 10\% improvement when trained on SNLI+MNLI. The best results are achieved by applying method \ref{ssec:direct} to either the first layer or all of the layers.


\subsection{Overall Comparison}
We add the Wordnet knowledge to all three models (Model I, Model II, BERT) and compare their performance on either clean or adversarial data. On SNLI dataset, for model I, we tune the number of layers $N \in \{1,2\}$, and the number of cross layers $N_{cr} \in \{1,2,3,4\}$. For model II, we tu $N$ from $\{1,2,3,4\}$. By the analysis in the previous section, we only apply the method \ref{ssec:direct} to $\{1\}$, $\{1,2\}$, $\{1,2,3\}$, or $\{1,2,3,4\}$ layers, and choose the best one on the SNLI validation set. On SNLI+MNLI task, we keep the same hyper-parameter settings as on SNLI. We report the model accuracy under all settings, and also the precision and recall for each class on the adversarial data. We append a \diamondmark to the model name if we modify some of its layers using the method \ref{ssec:direct} (e.g. BERT vs. BERT \diamondmark). The results are summarized in Table \ref{tab:all_models_comparison} and \ref{tab:all_models_comparison_mnli}.

\begin{table*}[htb]
\small
\centering
\caption{Comparison of all methods including model I, model II, and BERT. The models are trained on SNLI, and tested on both clean and adversarial SNLI data. The number of encoder layers is denoted by $N$, and the number of cross-encoder layers is $N_{cr}$. The precision and recall are reported in the order of [\textbf{entailment, neutral, contradiction}]. The \diamondmark models are those with modified multi-head attentions. The hyper-parameters are tuned on SNLI validation set.}
\label{tab:all_models_comparison}
\vspace{0.10in}
\begin{tabular}{cccccdadd}
\toprule
\multirow{2}{*}{\textbf{Method}} & \multirow{2}{*}{$\textbf{N}$} & \multirow{2}{*}{$\textbf{N}_{cr}$} & \textbf{Modified} & \textbf{No. of} & \textbf{Accuracy} & \textbf{Accuracy} & \textbf{Precision} & \textbf{Recall} \\
 & & & \textbf{layer(s)} & \textbf{parameters} & \textbf{(clean)} & \textbf{(adversarial)} & \textbf{(adversarial)} & \textbf{(adversarial)} \\
\toprule
Model I                          & 3 & 3 & none   & 12.2m   & 87.4\% &  
48.2\% & 32\%, 1\%, 99\% & 98\%, 26\%, 41\% \\
\midrule
Model I \diamondmark & 1 & 4 & 1,2,3  & 13.0m  & 88.5\% & \textbf{81.3\%} & 62\%, 1\%, 99\% & 92\%, 15\%, 80\% \\
\midrule
Model II                          &  4 & - & none  & 13m  & 82.2\% & 41.4\% & 37\%, 1\%, 98\% & 85\%, 53\%, 35\% \\
\midrule
Model II \diamondmark &  4 & - & 1,2,3,4 & 13m  & 82.5\% & 58.0\% & 44\%, 1\%, 98\% & 83\%, 45\%, 55\% \\
\midrule
BERT                             & 12 & - & none   & 108m & 90.5\% & 91.1\% & 74\%, 1\%, 99\% & 96\%, 11\%, 91\% \\
\midrule
BERT \diamondmark    & 12 & - & 1,2,3  & 108m & 90.1\% & \textbf{94.1\%} & 85\%, 4\%, 99\% & 97\%, 23\%, 94\% \\
\midrule
OpenAI GPT                       & 12 & - & none & 85m & 89.9\% & 83.8\% & - & - \\
\midrule
KIM     & - & - & - & - & 88.6\% & 83.5\% & - & - \\
\midrule
Wordnet & - & - & - & - & -      & 85.5\% & - & - \\
\midrule
Human   & - & - & - & - & 87.7\% & 94.1\% & - & - \\
\bottomrule
\end{tabular}
\end{table*}

\begin{table*}[htb]
\small
\centering
\caption{All the models are trained on SNLI+MNLI, and tested on clean MNLI and SNLI data, as well as adversarial SNLI data. The precision and recall are reported in the order of [\textbf{entailment, neutral, contradiction}]. The \diamondmark models are those with modified multi-head attentions. The hyper-parameters are set to be the same as in Table \ref{tab:all_models_comparison}.}
\label{tab:all_models_comparison_mnli}
\vspace{0.10in}
\begin{tabular}{ccccddadd}
\toprule
\multirow{2}{*}{\textbf{Method}} & \multirow{2}{*}{$\textbf{N}$} & \multirow{2}{*}{$\textbf{N}_{cr}$} & \textbf{Modified} & \textbf{Accuracy} & \textbf{Accuracy} & \textbf{Accuracy} & \textbf{Precision} & \textbf{Recall} \\
 & & & \textbf{layer(s)} & \textbf{(clean SNLI)} & \textbf{(clean MNLI)} & \textbf{(adversarial)} & \textbf{(adversarial)} & \textbf{(adversarial)} \\
\toprule
Model I                 &  3 & 3 & none  & 87.3\% & 72.8\% & 63.2\% & 53\%, 1\%, 99\% & 98\%, 34\%, 58\% \\
\midrule
Model I \diamondmark    &  1 & 4 & 1,2,3 & 87.1\% & 76.4\% & 84.4\% & 66\%, 1\%, 99\% & 98\%, 13\%, 83\% \\
\midrule
Model II                &  4 & - & none  & 82.7\% & 67.4\% & 54.7\% & 47\%, 1\%, 98\% & 88\%, 45\%, 50\% \\
\midrule
Model II \diamondmark   &  4 & - & 1,2,3 & 82.0\% & 68.2\% & 65.8\% & 47\%, 1\%, 99\% & 90\%, 36\%, 63\% \\
\midrule
BERT                    & 12 & - & none  & 90.2\% & 83.3\% & 93.2\% & 81\%, 2\%, 99\% & 99\%, 15\%, 93\% \\
\midrule
BERT \diamondmark       & 12 & - & 1,2,3 & 90.6\% & 83.0\% & \textbf{93.9\%} & 82\%, 3\%, 99\% & 99\%, 19\%, 94\% \\
\bottomrule
\end{tabular}
\end{table*}

For SNLI task, on the clean test data, BERT gives the best accuracy of 90.5\%, while Model I \diamondmark gets 88.5\%, slightly worse than 88.6\% by KIM. On the adversarial data, BERT \diamondmark gives the best result of 94.1\%, while Model I gets 81.3\% and KIM gets 83.5\%. However, our method is more straightforward as it only needs to be applied to multi-head attentions, and hence can be used in any model using attention mechanism (e.g. BERT). On the contrary, the way to add external knowledge in KIM is designed for the particular model, and cannot be directly transferred to other models.

Another observation is that despite the complexity of BERT, our method can further improve its accuracy on the adversarial data by 3\%. The recall on neutral class goes from 11\% to 23\%, and the precision on entailment examples increases from 74\% to 85\%. In fact, the 94.1\% accuracy equals to the estimated human performance reported in \citet{glockner2018breaking}.

However, both precision and recall on the neutral class are very low, indicating that the models are not good at predicting neutral labels, even with external knowledge. Because of the imbalance of the adversarial data set, the boost in accuracy mainly comes from predicting more contradiction instances correctly. For example, the recall of Model I on contradiction examples improves by 39\%, which is very close to the increase in overall accuracy.

\subsection{Comparision with Posthoc Modification}
\begin{table}[htb]
\small
\centering
\caption{Comparison of four scenarios on SNLI task. The \cmark mark means adding external knowledge in that phase, and \xmark means not adding any external knowledge. The Model I and Model I \diamondmark reported in Table \ref{tab:all_models_comparison} are used for comparison. For Model I, we add external knowledge to the first and second cross layers in scenario 4.}
\label{tab:training_inference}
\vspace{-0.10in}
\begin{tabular}{ccda}
\toprule
\textbf{Training} & \textbf{Inference} & \textbf{Accuracy} & \textbf{Accuracy} \\
\textbf{phase}    &   \textbf{phase}   & \textbf{(clean)}  & \textbf{(adversarial)}\\
\toprule
\cmark & \cmark & 88.5\% & 81.3\% \\
\midrule
\cmark & \xmark & 79.6\% & 57.2\% \\
\midrule
\midrule
\xmark & \xmark & 87.4\% & 48.2\% \\
\midrule
\xmark & \cmark & 84.6\% & 51.0\% \\
\bottomrule
\end{tabular}
\end{table}

In previous experiments, we apply the proposed method \ref{ssec:direct} to layers during both training and inference time. The reason is if we add external knowledge to one layer, the subsequent layers will learn to use this information. To show that training with external knowledge is crucial, we train another model normally but add the external knowledge in the inference time. The result is in Table \ref{tab:training_inference} together with three other scenarios.

The experiment results show that adding the external knowledge to the model in a posthoc manner (scenario 4) does not help improve the robustness. Without seeing the information during training, the model is unable to figure out how to incorporate it in the inference. On the other hand, from the first and second scenarios, we observe that dropping external knowledge in the inference will also lead to significant decrease in accuracy on the adversarial data. Therefore, it is keen to use the knowledge in both phases. 

%% file: subfiles/conclusion.tex
\section{Conclusion}

We proposed to make NLI models more robust by incorporating external lexical knowledge into the attention mechanism. This method can be widely used in any model that has an attention component. To test the effectiveness of this approach, we apply it to two popular types of NLI models, and observe significant accuracy boost on the adversarial SNLI data set. Finally, even for a gigantic model like BERT, our method still provides extra improvement on the robustness, and achieves the human-level accuracy on the SNLI adversarial test set.

%% file: subfiles/supplement.tex
\section{Supplemental Material}
\label{sec:supplemental}


\begin{table*}[htb]
\small
\centering
\caption{Accuracy of \textbf{Model I} trained on \textbf{SNLI+MNLI} and tested on clean SNLI/MNLI and adversarial SNLI data. We set the parameter $b=10$. The modified layers are the layers where we add external knowledge using method \ref{ssec:direct}. The absolute increase is the improvement of accuracy compared to the corresponding baseline model.}
\label{tab:model_I_direct_mnli}
\vspace{-0.00in}
\begin{tabular}{cccc|ccdc}
\toprule
\multirow{2}{*}{$\textbf{L}$} & \multirow{2}{*}{$\textbf{L}_{cr}$} & \textbf{Modified} & \textbf{No. of} & \textbf{Accuracy} & \textbf{Accuracy} & \textbf{Accuracy} & \textbf{Absolute} \\
& & \textbf{layer(s)} & \textbf{parameters} & \textbf{(clean SNLI)} & \textbf{(clean MNLIm)} & \textbf{(adversarial)} & \textbf{increase} \\
\midrule
1 & 1 & none & 6.8m & 86.5\% & 72.5\% & 63.2\% & baseline \\
\midrule
1 & 1 & 1    & 6.8m & 86.8\% & 73.2\% & 83.1\% & 20.9\% \\
\midrule
1 & 2 & none & 8.9m & 87.3\% & 72.1\% & 62.9\% & baseline \\
\midrule
1 & 2 & 1    & 8.9m & \textbf{88.7\%} & \textbf{76.1\%} & \textbf{85.2\%} & 22.3\% \\
\midrule
1 & 2 & 2    & 8.9m & 87.7\% & 74.6\% & 84.0\% & 21.1\% \\
\midrule
1 & 2 & 1,2  & 8.9m & 88.2\% & 74.9\% & 84.6\% & 21.7\% \\
\bottomrule
\end{tabular}
\vspace{-0.30in}
\end{table*}


\begin{table*}[htb]
\small
\centering
\caption{Accuracy of \textbf{Model II} trained on \textbf{SNLI+MNLI} and tested on clean SNLI/MNLI and adversarial SNLI data. We set the parameter $b=10$. The modified layers are the layers where we add external knowledge using method \ref{ssec:direct}. The absolute increase is the improvement of accuracy compared to the corresponding baseline model.}
\label{tab:model_II_direct_mnli}
\vspace{-0.00in}
\begin{tabular}{ccc|ccdc}
\toprule
\multirow{2}{*}{$\textbf{L}$} & \textbf{Modified} & \textbf{No. of} & \textbf{Accuracy} & \textbf{Accuracy} & \textbf{Accuracy} & \textbf{Absolute} \\
 & \textbf{layer(s)} & \textbf{parameters} & \textbf{(clean SNLI)} & \textbf{(clean MNLIm)} & \textbf{(adversarial)} & \textbf{increase} \\
\midrule
3 & none  & 11.0m & 81.8\% & 66.7\% & 54.3\% & baseline \\
\midrule
3 & 1     & 11.0m & 82.4\% & 67.8\% & 65.6\% & 11.3\% \\
\midrule
3 & 2     & 11.0m & 82.2\% & 66.9\% & 64.5\% & 10.2\% \\
\midrule
3 & 3     & 11.0m & 81.9\% & 67.5\% & 63.8\% & 9.5\% \\
\midrule
3 & 1,2,3 & 11.0m & 81.7\% & 67.2\% & 65.2\% & 10.9\% \\
\bottomrule
\end{tabular}
\end{table*}